# Iterative Augmentation with Summarization Refinement (IASR) Evaluation for Unstructured Survey data Modeling and Analysis

Payal Bhattad, Sai Manoj Pudukotai Dinakarrao and Anju Gupta*

*Abstract*— Text data augmentation is a widely used strategy for mitigating data sparsity in natural language processing (NLP), particularly in low-resource settings where limited samples hinder effective semantic modeling. While augmentation can improve input diversity and downstream interpretability, existing techniques often lack mechanisms to ensure semantic preservation during large-scale or iterative generation, leading to redundancy and instability. This work introduces a principled evaluation framework for large language model (LLM) based text augmentation, comprising two components: (1) Scalability Analysis, which measures semantic consistency as augmentation volume increases, and (2) Iterative Augmentation with Summarization Refinement (IASR), which evaluates semantic drift across recursive paraphrasing cycles. Empirical evaluations across state-of-the-art LLMs show that GPT-3.5 Turbo achieved the best balance of semantic fidelity, diversity, and generation efficiency. Applied to a real-world topic modeling task using BERTopic with GPT-enhanced few-shot labeling, the proposed approach results in a 400% increase in topic granularity and complete elimination of topic overlaps. These findings validated the utility of the proposed frameworks for structured evaluation of LLM-based augmentation in practical NLP pipelines.

*Index Terms*- Large Language Models (LLMs), Text Data Augmentation, Topic Modeling, Semantic Similarity, Natural Language Processing (NLP)

This work was supported in part by the U.S. National Science Foundation under Grant 2305245. *Corresponding author: Anju Gupta*.

Payal Bhattad is with Electrical Engineering and Computer Science Department at the University of Toledo, Toledo, OH 43606, USA (e-mail: payal.bhattad@rockets.utoledo.edu).
Sai Manoj Pudukotai Dinakarrao is Electrical and Computer Engineering Department at George Mason University, Fairfax, VA 22030, USA (e-mail: spudukot@gmu.edu).
Anju Gupta is with Mechanical, Industrial and Manufacturing Engineering Department, at the University of Toledo, Toledo, OH 43606, USA (anju.gupta@utoledo.edu).

## I. INTRODUCTION

High-quality, domain-specific text data is fundamental to the success of natural language processing (NLP) systems [1], underpinning applications such as sentiment analysis [2], classification [3], and topic modeling [4] across diverse domains including social media, product reviews, academic literature, and organizational documents. Despite rapid advances in NLP, many real-world datasets, particularly those derived from open-ended surveys, short stories, or specialized corpora, remain sparse, disjointed, or contextually limited. This data scarcity poses a critical challenge, limiting NLP models' ability to learn robust semantic representations and capture meaningful patterns. The issue is especially acute for user-generated or domain-specific content, such as open-ended survey responses, customer feedback, or incident reports, where limited data volume impedes the discovery of latent themes and interpretable insights. As a result, the success of many NLP tasks continues to rely heavily on the availability of sufficient, high-quality training data, highlighting the ongoing need for effective strategies to overcome data sparsity in both mainstream and specialized applications [5].

This study examines the unstructured component of real-world datasets, specifically open-ended textual responses, which frequently lack the volume necessary for reliable semantic modeling and meaningful insight generation. Although source datasets may include both structured and unstructured data, this work's evaluation focuses on the quality of augmentation applied to the free-text component. Topic modeling, the most widely used technique for uncovering hidden topical patterns in large corpora, often operates suboptimally in these settings [6]. Traditional models such as Latent Dirichlet Allocation (LDA) tend to produce overly coarse topics in resource-constrained environments and fail to capture subtle distinctions between closely related concepts [7]. These limitations diminish interpretability and reduce the practical relevance of topic modeling in real-world applications [8, 9].

Data augmentation has emerged as a practical solution to address the data sparsity problem by synthetically expanding corpora while maintaining the semantic integrity of original texts [10]. In the context of NLP, this process involves generating paraphrased or modified versions of input data to enhance model generalizability without requiring the collection of additional real-world data [11]. However, conventional augmentation techniques such as synonym replacement, random deletion, and back-translation often fail to provide the necessary lexical diversity or semantic fidelity required for nuanced tasks like topic modeling, motivating the use of context-aware generative models [12, 13]. For instance, fastText [14] and Global Vectors (GloVe) embeddings [15], while preserving core semantics, exhibit limited lexical variation, with n-gram diversity scores of just 0.096 and 0.097 in our preliminary evaluation. On the other hand, back-translation introduces uncontrolled semantic drift, which can potentially distort the thematic structures of the corpus [16]. These findings suggest that traditional augmentation methods either introduce noise or lack diversity, making them suboptimal for tasks that rely on thematic consistency and interpretability.

To overcome the limitations of conventional augmentation techniques, this study leverages the potential of context-aware generative models, particularly large language models (LLMs), building upon earlier transformer-based architectures such as Bidirectional Encoder Representations from Transformers (BERT) [40] and T5 [43], as an advanced alternative for generating diverse yet semantically consistent text. LLMs such as GPT-3.5, Generative Pre-trained Transformer (GPT-4), and Claude Sonnet 3.5 are generative Artificial Intelligence (AI) models trained on vast corpora of internet-scale text, enabling them to generate human-like responses with high contextual fidelity [17, 18]. While earlier studies have primarily leveraged LLMs for classification or question-answering tasks, this study uniquely applies their generative capacity to augment textual data for improved topic modeling performance, a domain that remains underexplored in the augmentation literature. This shift in generative capabilities marks a significant departure from traditional methods, as several studies have demonstrated that LLMs can produce contextually relevant content that is coherent and diverse [19, 20]. Their transformer-based architecture enables deep semantic understanding, making them particularly effective at generating paraphrased outputs that retain the original intent while introducing linguistic variation. Unlike rule-based or embedding-driven methods that struggle to maintain a balance between diversity and semantic retention, LLMs provide scalable augmentation capabilities across varied linguistic structures. Additionally, recent studies [21-23] have shown that LLM-generated augmentations outperform traditional methods in classification and sentiment analysis tasks.

Despite growing interest in using LLMs for data augmentation in classification or sentiment analysis [24, 25], a critical research gap remains: comparatively few have conducted systematic comparisons across LLMs for augmentation quality or controlled for their downstream impact on topic modeling using quantifiable and replicable metrics. Much of prior work relies upon a single, unexplained model or qualitative measures [26, 27]. Traditional evaluation methods such as Bilingual Evaluation Understudy (BLEU) [28] and Recall-Oriented Understudy for Gisting Evaluation (ROUGE) [29] primarily measure surface-level lexical overlap, failing to capture semantic equivalence when paraphrased expressions retain the same meaning. A more robust evaluation using contextual embeddings, such as cosine similarity with Sentence-BERT [30], offers better semantic alignment but is often employed in static contexts, without accounting for scale or iterative stability [31]. In contrast, this study introduces dynamic evaluation settings that extend beyond static similarity, assessing LLM output stability under both volume scaling and iterative transformations, an area overlooked by prior augmentation literature.

Building upon this gap, existing research on iterative refinement, such as work by [32] and [33], is largely confined to summarization tasks, leaving the assessment of augmentation quality through repeated refinement cycles underexplored. This research addresses this void directly by outlining two novel assessment approaches: Scalability Analysis, which tests semantic consistency under varying augmentation volumes, and IASR (Iterative Augmentation with Summarization Refinement), which evaluates the stability and content fidelity of augmentations under a series of refinement iterations. Both frameworks utilize cosine similarity with context-aware embeddings to provide robust, model-agnostic assessment pipelines. *To the best of our knowledge, this is the first work to design augmentation evaluation strategies tailored to the needs of topic modeling under data scarcity, rather than adapting metrics designed for classification or summarization tasks.*

The key contributions of this work could be outlined in a threefold manner, as follows:

1) This work introduces two novel evaluation frameworks for LLM-based text augmentation: Scalability Analysis, which quantifies semantic preservation across increasing augmentation volumes, and Iterative Augmentation with Summarization Refinement (IASR), which evaluates semantic stability through successive iterations.
2) A systematic comparison of four state-of-the-art LLMs for text augmentation reveals GPT-3.5 Turbo as providing the optimal balance between semantic preservation of 0.855 similarity and computational efficiency, requiring only 16 minutes to generate 100 augmentations per sentence, 40% faster than Claude 3.5 Sonnet, which took 28 minutes for the same task.
3) A hybrid pipeline that combines LLM-driven augmentation with BERTopic and few-shot GPT topic labeling, achieving a 400% increase in topic granularity, with coherence of 0.526 preserved and topic overlap eliminated.

This study systematically benchmarks four state-of-the-art large language models using the proposed evaluation methodologies to assess the effectiveness of augmentation in low-resource NLP settings. To ensure practical applicability, the framework is applied to open-ended textual responses from the Faculty Career-Life Survey. The highest-performing model is then used to augment this real-world dataset, with subsequent topic modeling conducted using BERTopic and a few-shot GPT-based topic labeling approach. An ablation study further isolated the individual and combined effects of augmentation and GPT-based structuring, offering detailed insights into how LLM-driven augmentation influences thematic granularity and coherence. This controlled experimental design distinguishes the work from previous studies by providing ablation-based evidence of the specific impact of augmentation on downstream interpretability.

## II. RELATED WORK

### A. Traditional Data Augmentation Techniques

Data augmentation in natural language processing refers to the generation of synthetic textual samples by applying transformations to the original data [34]. It is frequently employed to mitigate data scarcity and improve generalization in downstream NLP tasks. Early augmentation methods can be categorized based on the granularity of the applied transformations, including character, word, and sentence levels [13].

At the character level, techniques such as random character insertion, deletion, or swap are used to simulate typographical errors, thereby enhancing model robustness to noisy input [35]. Word-level augmentation introduces lexical variation through approaches like synonym replacement [13] and embedding-based substitution. Lexical resources, such as WordNet [36], have been employed for synonym substitution. At the same time, Word to Vector (Word2Vec) and GloVe embeddings facilitate the replacement of words with the top-k most similar tokens in the vector space [15, 37]. However, proximity in embedding space does not always correspond to actual semantic similarity. For instance, the words "hot" and "cold" may appear in similar contexts but possess opposite meanings, potentially introducing semantic distortion [38]. This problem has led to the development of counter-fitting techniques that refine embeddings by minimizing the distance between synonyms and maximizing it between antonyms [39].

Contextual augmentation methods such as BERT-based substitution have addressed these shortcomings by generating replacements based on bidirectional context [40]. Instead of relying on static similarity, masked language models predict contextually appropriate tokens, enabling more coherent and semantically faithful augmentations. Sentence-level techniques provide higher-level transformation by generating paraphrased variations of complete sentences. Back translation is one such method, where the input text is translated into an intermediate language and then back to the original, creating diverse phrasing while retaining the core meaning [41]. Although back translation introduces linguistic diversity, prior work has reported semantic drift and inconsistencies, especially in complex or domain-specific content [12, 16]. While traditional augmentation methods have contributed to improved data diversity, they often struggle to preserve semantic integrity [26] or generate nuanced variations necessary for tasks such as topic modeling. These limitations provide the rationale for exploring more advanced, context-aware augmentation methods using large language models.

### B. LLM-based Data Augmentation

Pretrained transformer-based language models (PLMs) have revolutionized the field of natural language processing by enabling contextualization at deeper levels and highly flexible language generation [42]. PLMs can typically be classified into three categories: autoregressive models such as GPT, masked language models like BERT, and encoder-decoder models like T5 [40, 43]. Traditional PLMs typically contain 100 million to 1 billion parameters and are typically fine-tuned on downstream tasks to achieve competitive performance on text classification, sentiment analysis, and summarization.

New advancements in context-aware generative models, including increased model capacity and refined training objectives, have enabled the development of LLMs such as GPT-3.5, GPT-4, and Claude Sonnet. [18]. Such models, which have been trained on internet-scale corpora, possess hundreds of billions of parameters and employ an autoregressive generation to produce fluent and semantically dense text. LLMs are trained to learn context-sensitive and domain-adaptive latent text representations. For example, the semantic value of the word "assessment" in context to the present study might be radically variable depending on whether it is being used in the context of academic evaluation, faculty promotion, or student performance, demonstrating the relevance of contextual modeling for augmentation processes. Unlike small PLMs, which often require extensive fine-tuning, LLMs are generalizable to a wide range of tasks without having to be retrained [44].

A distinguishing feature of models like ChatGPT is their reinforcement learning with human feedback (RLHF) training, a procedure that aligns model output with human preference and improves generation quality [17, 45]. This architecture enables LLMs to produce improved text that maintains syntactic fluency, semantic fidelity, and stylistic coherence. Some studies have indicated that augmentations generated by LLMs are superior to traditional approaches in classification and sentiment tasks, offering greater diversity without compromising meaning [21, 46]. However, these studies have not systematically examined their reliability for topic modeling or their stability under iterative refinement, both of which are evaluated in this work. This study follows progress and evaluates the augmentation capability of four LLMs, GPT-3.5 Turbo, GPT-4 Turbo, GPT-4-o-mini, and Claude 3.5 Sonnet, in terms of contextual consistency, lexical diversity, and computational efficiency. By contrasting how well they did in generating syntactically variable but semantically correct augmentations, this research improves the understanding of LLM application in few-shot, domain-specific tasks such as topic modeling.

### C. Evaluation of Augmentation and Topic Modeling Impact

Evaluating the quality of text augmentation is essential for ensuring that synthetic samples are both semantically faithful and contextually diverse. Traditional evaluation metrics, such as BLEU [28] and ROUGE [29], focus on surface-level lexical overlap, making them insufficient for assessing paraphrastic variations where lexical choices differ but the semantic intent is preserved. More recent approaches leverage embedding-based similarity measures, such as cosine similarity computed over Sentence-BERT or other contextual encoders, to capture deeper semantic alignment [30, 31].

Despite these advances, most existing evaluations are task-specific and static, offering limited insight into the extent to which augmentation quality generalizes or persists with iterative refinement. [32] and [33] explored stability in summarization pipelines but did not extend their results to more general augmentation contexts. To address these limitations, this study introduces two model-agnostic evaluation frameworks for LLM assessment: Scalability Analysis and IASR. Unlike existing evaluations that often assess augmentation using static, task-specific metrics [30, 31], these frameworks offer dynamic diagnostics that reflect semantic stability across scale and iterative refinement. Scalability assesses how semantic similarity between original and generated text changes with the quantity of generated samples, thereby evaluating consistency at scale. IASR evaluates whether augmentation quality is preserved across multiple rounds of iterative refinement, simulating real-world generation chains in which text is repeatedly paraphrased. Both models utilize cosine similarity with BERT-based embeddings as a robust and reproducible metric for semantic comparison.

Aside from quality augmentation, its use in topic modeling remains less explored. Topic modeling has generally relied on statistical approaches such as LDA [7], which in low-data settings tend to provide diffuse or meaningless topics. Topic coherence refers to the degree of semantic similarity among the top keywords within a topic. It is widely used to assess how interpretable or meaningful the identified topics are to humans, with higher coherence values indicating better thematic alignment [47].

Neural models such as BERTopic have improved topic coherence and accuracy [48], but their performance remains highly sensitive to input quantity and quality. Early research by [49] highlights the connection between dataset size and topic coherence, illustrating how sparsity reduces interpretability. While some research, such as [50], brought pseudo-document simulations to facilitate topic discovery, these are found on statistical synthesis rather than semantically grounded generation. Equivalently, [51] augmentation studies have mostly focused on classification rather than interpretability-driven topic extraction. This study bridges that gap by directly linking augmentation quality to topic modeling performance through coherence, diversity, and structure-based evaluation.

III. METHODOLOGY

*A. Experimental Overview and LLM Evaluation Setup*

The methodology of this study was designed to systematically evaluate LLM-based data augmentation and its impact on topic modeling performance in low-data NLP settings. The dataset used in this study consisted of real-world responses collected through an organizational feedback survey, consisting of a combination of numerical ratings, Likert-scale items, and open-ended textual responses. This study primarily focused on the qualitative component to explore augmentation and topic modeling techniques in a data-scarce environment where limited text availability challenges effective theme extraction and interpretability. All experiments were conducted on a hybrid computing environment comprising Google Colab Pro+ and institutional high-performance computing (HPC) infrastructure. Augmentation models were accessed via API endpoints for GPT-3.5 Turbo, GPT-4 Turbo, GPT-4-o-mini, and Claude 3.5 Sonnet. Hardware configurations included NVIDIA A100 (40 GB VRAM) and Tesla T4 (16 GB VRAM) GPUs. The implementation stack featured Python 3.10 or later, Hugging Face Transformers, BERTopic, Sentence-Transformers, Scikit-learn, and NLTK. To ensure uniformity in document representation across all experimental configurations, all models were evaluated using the same sentence embedding model, *all-MiniLM-L6-v2* [52].

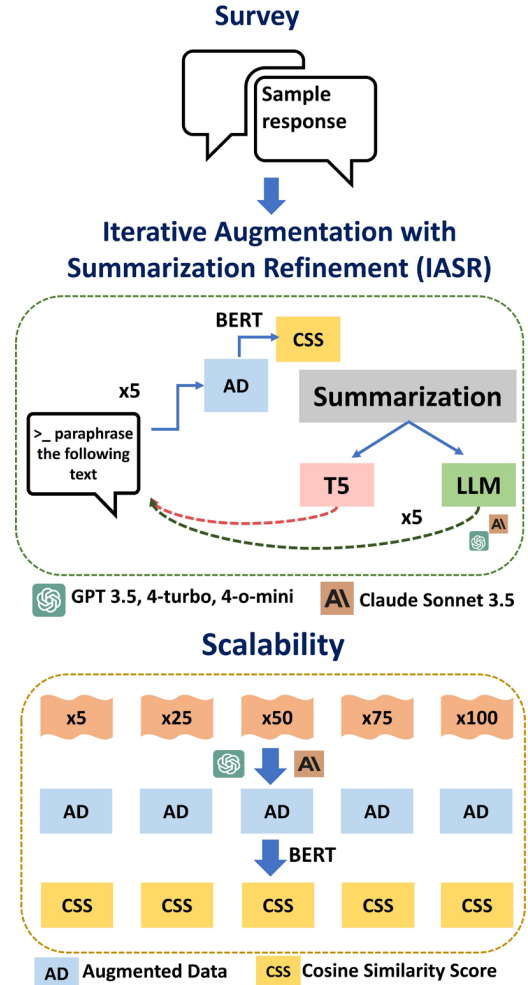

**Fig. 1**. Overview of the proposed workflow combining Iterative Augmentation with Summarization Refinement (IASR) and Scalability Analysis for enhancing topic modeling of low-resource survey data.

The process involved two main components. First, it evaluates augmentation methods by measuring their quality using cosine similarity, analyzing the impact of scalability, and introducing a novel evaluation technique called Iterative Augmentation with Summarization Refinement (IASR). **Fig.1.** illustrates the overall workflow used to evaluate both IASR and Scalability methodologies. In the IASR pipeline (top panel), raw survey responses were paraphrased LLMs such as GPT-3.5 and Claude Sonnet 3.5 to generate augmented data (AD). These paraphrased responses were encoded using BERT, and their semantic consistency is evaluated via cosine similarity scores (CSS). To mitigate semantic drift, the augmented corpus is refined through multi-stage summarization using models like T5 and additional LLMs. In the Scalability module (bottom panel), varying augmentation scales (5×–100×) are evaluated to assess their impact on topic model coherence and separation. Cosine similarity metrics guide fidelity assessment, enabling optimal augmentation depth selection. This dual-pronged framework enables systematic benchmarking of LLM-based augmentation strategies for unsupervised text mining. Four state-of-the-art context-aware generative models GPT-3.5 Turbo, GPT-4 Turbo, GPT-4-o Mini, and Claude Sonnet 3.5, were evaluated for their effectiveness in text augmentation. These models were selected based on their performance in natural language generation (NLG), contextual coherence, and computational efficiency [52-54]. GPT-3.5 Turbo and GPT-4 Turbo were chosen for their strong contextual understanding and consistent semantic generation. GPT-4-o Mini was included to assess performance-efficiency trade-offs, while Claude Sonnet 3.5 introduced architectural diversity for cross model comparison. To assess the quality of augmented text, the study employed cosine similarity with BERT embeddings, a widely adopted metric in NLP for measuring semantic alignment between textual representations. Unlike n-gram-based evaluation metrics such as BLEU [28] or ROUGE [29], which rely on exact token overlap, cosine similarity evaluates contextual proximity, facilitating the assessment of LLM-generated paraphrases effectively. For embedding generation, sentence-transformers/bert-base-nli-mean-tokens [30], a BERT-based model optimized for sentence similarity tasks, was utilized. Each original response and its corresponding augmented text were converted into 768-dimensional embeddings, with their semantic similarity [55]. A higher cosine similarity score indicates greater semantic retention, while a lower standard deviation across augmentations suggested increased consistency in the quality of generation. The selection of cosine similarity over conventional evaluation metrics was motivated by its effectiveness in capturing contextual relationships in text augmentation. Traditional n-gram-based metrics such as BLEU [28] and ROUGE [29] primarily measure lexical overlap, which does not adequately assess semantic preservation in LLM-generated text, where rephrasing and structural variations are common. In contrast, BERT-based embeddings represent text in a high-dimensional semantic space, enabling a more precise and reliable assessment of meaning retention in augmented responses [56]. Furthermore, cosine similarity was computationally efficient and scalable, making it well-suited for evaluating augmentation methods across varying dataset sizes [57] that further ensured that LLM-generated augmentations maintain semantic integrity and do not introduce semantic drift as augmentation volume increases.

**Fig.2.** illustrates the workflow of computing semantic similarity between original and augmented survey responses using BERT-based sentence embeddings. The input text is first tokenized and encoded via a pretrained BERT encoder to produce contextualized sentence embeddings. Three embedding strategies are compared: mean of the last hidden layer, [CLS] token vector, and SBERT-WK subspace analysis. Cosine similarity between sentence embeddings u and v is calculated as: $cos\theta = \frac{u \cdot v}{\|u\|\|v\|}$ , ], where $u$ and $v$ represent the sentence embeddings of the original and augmented text, respectively. quantifying the semantic alignment between original and augmented responses. Sentence representations were derived using the *bert-base-nli-mean-tokens* model, and under various augmentation techniques including synonym substitution (WordNet), static embeddings Word2Vec, GloVe, contextual LMs BERT, back-translation, and generative models such as GPT-3.5 Turbo. In addition to quantitative similarity scores, qualitative assessments were conducted to evaluate semantic fidelity and fluency across augmentation methods.

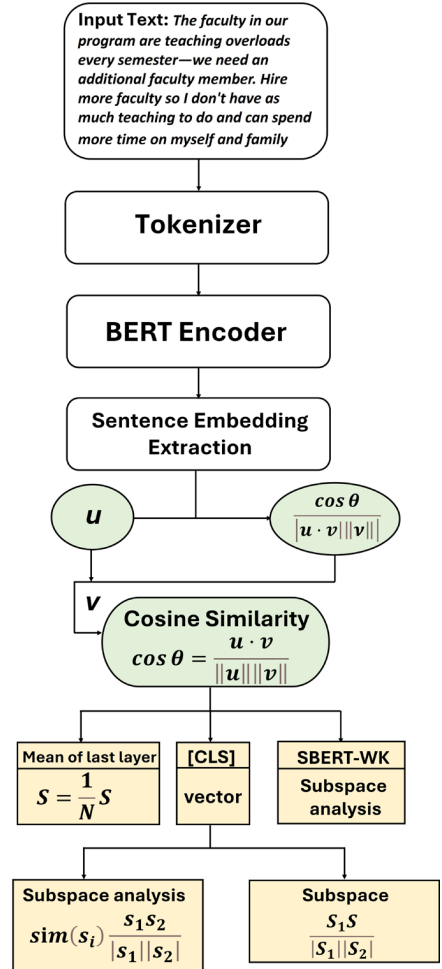

**Fig.2.** Workflow for computing semantic similarity between original and augmented survey responses using BERT-based sentence embeddings.

TABLE I provides a summary of traditional methods resulting in either syntactic incoherence or semantic distortion. To qualitatively assess the fidelity and coherence of different text augmentation strategies, a representative faculty survey response was subjected to various augmentation methods, including traditional word embedding–based techniques Word2Vec, GloVe, BERT, lexical substitution via WordNet, back-translation and prompt-based paraphrasing using GPT-3.5 Turbo. Each method was applied to the same input sentence emphasizing faculty recognition and merit raises. The results revealed distinct performance profiles. Word2Vec and GloVe, which rely on static word embeddings, often generate syntactically flawed or semantically inconsistent outputs due to out-of-context substitutions.

TABLE I
Comparison of augmentation methods applied to a representative faculty survey response using both traditional and large language model (LLM)-based techniques.

| Method | Description | Augmented Output |
|---|---|---|
| Original | | In general, the university is not good at recognition of faculty efforts, particularly those of new faculty. Merit raises need to be returned. Merit raises need to return. |
| Word2vec | Substitute or insert word randomly using word embeddings similarity. | On general the university is not loud_bang_Timms Banqueting_Suite prominence of faculty efforts, particularly those of new faculty. OOM escalates must to back. Merit raises wanted to reclaim. |
| GloVe | Substitute or insert word randomly using word embeddings similarity. | In office the theology is not good around recognition of harvard peace, particularly those of new cornell. Merit stirs need likely left. Merit raises could return. |
| BERT | Substitute or insert words by contextual word embeddings. | university was not looking at recognition of faculty efforts, particularly those of new members. merit raises reluctance to graduate. Honor raises need before return. |
| WordNet | Substitute or insert word by synonym. | In general university is not ripe at acknowledgment of faculty try, especially those of new faculty. Meritoriousness raises the need to return. Deservingness raises necessity to return. |
| Back-Translation (en-de/de-en) | Substitute or insert word using back translation. | In general, the University is not good at recognizing the efforts of the faculties, especially those of the new faculties. Merits increase the need to return. Merits increase the need to return. |
| GPT-3.5 Turbo | Prompt: "Paraphrase the following text:" | Overall, the university lacks in acknowledging the efforts of faculty members, especially new ones. It is important for merit raises to be reinstated. |

BERT-based augmentation showed improved contextual fluency but sometimes introduced thematic shifts or less relevant terms. WordNet-based synonym replacement maintained grammaticality but altered the tone and nuance of the original statement. Back-translation preserved sentence structure more faithfully but introduced minor lexical changes.

Notably, GPT-3.5 Turbo, guided by a simple paraphrasing prompt, produced coherent and semantically aligned variants that preserved the original message while enhancing clarity and readability. This validated the need for context-aware generative models in semantically demanding applications such as topic modeling.

*B. Evaluation Technique: Scalability and IASR*

The evaluation of LLM-based text augmentation was conducted through two complementary methodologies: Scalability Analysis, which assessed the impact of augmentation volume on semantic consistency, and IASR, the proposed novel diagnostic framework designed to assess quality under iterative refinement. These methodologies jointly provide a multifaceted evaluation of augmentation robustness, as illustrated previously in Fig. 1.

For scalability analysis, two representative responses were selected from the dataset to reflect different levels of linguistic complexity. The first was a concise imperative statement, "*The faculty in our program are teaching overloads every semester, we need an additional faculty member. Hire more faculty so I don't have as much teaching to do and can spend more time on myself and family.*". The second was a long-form, multi-thematic narrative highlighting structural and emotional concerns, "*The one program I felt was making a difference on campus was canceled and almost taboo to be discussed. Our department is stretched too thin and at times I feel that I am the only one working toward students' success and belonging. I cannot even get colleagues to respond to emails at times to discuss what would be helpful to them. Teaching and committee work seem to fall on lecturers more and more; however, we are paid the least and often feel unvalued by colleagues. Students and faculty want the ability to work from home and face-to-face. A hybrid approach would benefit all. There is no official training for faculty on best practices for teaching. Without the administration backing up this type of training, it will never happen, and students will continue to feel undervalued in the classrooms. The pay for lecturers. Focus on student success and best practices in the classroom.*". These contrasting examples enabled testing of semantic retention across simple and complex textual inputs.

For each sentence, augmentations were generated at five scales: 5x, 25x, 50x, 75x, and 100x. Semantic similarity between original and augmented responses was computed using cosine similarity with BERT embeddings. Three quantitative measures were used to evaluate augmentation stability: (1) Semantic Retention: The mean cosine similarity indicated how well the original meaning was preserved at each augmentation level. (2) Consistency Across Augmentations: The standard deviation of similarity scores measured generation stability, with lower variance implying more consistent output quality. Together, these metrics provided a structured understanding of how LLM-generated augmentations scale with quantity and whether semantic integrity is preserved under high-generation scenarios.

While Scalability Analysis quantified how augmentation quality changed with volume, it did not address whether incorporating structured refinement steps can improve semantic fidelity. In practical NLP pipelines such as document generation, chat systems, and text generation, text is often developed iteratively, where one output relies on previous responses rather than being generated anew. This recursive mechanism jeopardizes cumulative semantic drift, where minor shifts in meaning accumulate over multiple iterations, reducing contextual accuracy and clarity. Prior work has demonstrated that Iterative summarization can improve output quality across subsequent rounds. For instance, the SummIt framework [32] achieved a 38% reduction in factual errors and a ~12-15% improvement in ROUGE scores after three summarization cycles. These results suggested that multi-step summarization is a good quality-control process, enhancing semantic coherence, factual consistency, and linguistic clarity. However, although iterative summarization has been explored in summary generation, its use in fine-tuning LLM-based data augmentation pipelines is limited. To address this gap, IASR is presented as a new diagnostic testbed for augmentation robustness. This evaluation framework is designed to assess how well LLMs preserve content semantics and generation consistency across repeated refinement cycles. IASR tests programmatically whether including summarization steps in the augmentation pipeline of an LLM help to maintain the intrinsic meaning of the text and enhance controlled lexical variability with iterations.

IASR integrated summarization into augmentation cycles and evaluates how LLMs respond to recursive summarization steps. Summarization condenses augmented responses into concise forms at every cycle, which are re-augmented, simulating refinement chains in real-world settings. This allowed for measurement of whether semantic fidelity is preserved or lost across cycles. The same semantic similarity metrics, mean cosine similarity and standard deviation of similarity scores, were tracked across refinement rounds to determine whether summarization contributed to controlled variability or introduced semantic drift. IASR was implemented in two configurations, as shown in **Fig. 3**: (1) External Summarizer (T5-Base): In this variation, Summarization and augmentation are performed by separate models. This prevents self-reinforcing biases and offers a neutral assessment of how summarization affects meaning retention. (2) Self-Summarization (Same LLM): In this, the same model used for augmentation performs summarization, allowing for the evaluation of whether an LLM can refine its own outputs without external guidance. This variation highlights the risks of recursive self-generation, including the compounding of stylistic repetition and semantic drift.

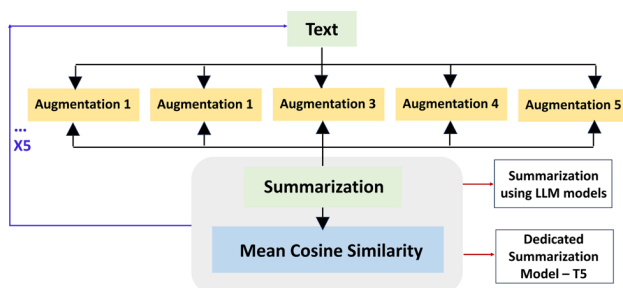

**Fig. 3.** Iterative Augmentation with Summarization Refinement (IASR) Workflow

This multi-stage IASR procedure began with the generation of multiple paraphrased variants of an input text using large language models (LLMs). In each iteration, a set of augmented responses is produced and subsequently summarized using either a dedicated summarization model such as T5-Base or an LLM-based summarizer. The resulting summaries distilled key semantic content from the augmented variants and serve as updated prompts for the next augmentation round. This summarize–augment cycle is repeated for a typical five iterations, enabling progressive refinement and semantic reinforcement. To evaluate semantic stability across iterations, the mean cosine similarity between the input and successive augmented outputs was computed.

In the self-summarization variant, the same LLM takes its own augmented output and generates a summary of it, which is then used as the basis for the next augmentation cycle, instead of using T5. In either case, the core idea is that the summarization step refines the content by compressing it, potentially eliminating redundancies or minor errors, before the next augmentation step expands it again. This iterative refinement continued for several cycles, producing a sequence of augmented texts that have been periodically distilled and regenerated. At each iteration, embedding-based cosine similarity metrics are computed between the latest augmented text and the original input, or an initial reference, to assess semantic preservation. Consistently high cosine similarity across iterations would indicate that the summarization steps are successfully preserving the core meaning through each regeneration cycle. Additionally, by comparing the diversity of phrasing in outputs, IASR monitored that the model still introduced some lexical variety preventing identical repetitions without altering the intended meaning. High similarity across iterations indicates that summarization helps maintain contextual fidelity, while stable variance across outputs suggests consistent generation. Comparing both IASR variants enables the evaluation of trade-offs between architectural independence (T5-Base) and adaptive self-correction, same LLM.

*C. Application to Topic Modeling and Impact Analysis*

To evaluate the effect of augmentation on thematic coherence, topic modeling using BERTopic was performed on augmented and non-augmented datasets. The most computationally efficient LLM, as established through a balance of both high mean cosine similarity and qualitative computational efficiency, was used in generating augmented samples. Although exact latency and cost measures were not tracked, model efficiency was estimated through responsiveness and ease of use when calling the Application Programming Interface (API). All text records were preprocessed with common natural language processing steps, such as lowercasing, punctuation removal, and stopword filtering with the Natural Language Toolkit (NLTK) corpus, to provide uniform formatting before topic extraction. The responses were then projected into dense vector representations using the all-MiniLM-L6-v2 Sentence Transformer model, chosen for its balance of computational efficiency and semantic accuracy.

BERTopic was applied to these embeddings to extract coherent topic clusters from both datasets. To systematically compare the quality of topics from original and augmented data, five metrics were computed. Topic coherence evaluated the semantic alignment between topic embeddings and their document embeddings. Topic diversity captured lexical richness by evaluating the ratio of unique terms across topics. Topic overlap, as formulated by the Jaccard similarity [59], was evaluated to assess lexical redundancy among topics. Intra-topic similarity quantified semantic cohesion within the topics, and inter-topic distance quantified the distinctiveness between topic centroids in the embedding space.

Although BERTopic was able to identify topic clusters, the resulting keyword-based labels were often redundant, ambiguous, or difficult to interpret, especially in terms of domain-specific feedback. To enhance the quality of topic labels, a GPT-based few-shot learning method was used as a post-processing step. Some topic clusters were also manually annotated by domain specialists to generate high-quality exemplars, which were used to guide GPT-3.5 Turbo in generating more accurate and comprehensible topic labels. The prompt design was oriented toward brevity and domain specificity, with label lengths constrained to two to five words. To reduce vagueness and ensure clarity, the prompt discouraged generic phrasing and encouraged the use of structured naming conventions. For example, "Administration: Leadership and Support" was preferred over vague alternatives such as "Leadership issues" or "Problems with admin," which lack thematic precision. This human-in-the-loop approach ensured that resulting topic labels accurately represented respondent concerns. To further assess the value of augmentation and refinement, an ablation study was conducted under three conditions: (1) topic modeling without augmentation, (2) topic modeling without GPT-based refinement, and (3) topic modeling with both augmentation and GPT-assisted topic labeling. This design enabled the isolation and measurement of individual and combined effects on topic interpretability and structure.

## IV. RESULT AND DISCUSSION

### A. Evaluation of Augmentation Quality: Scalability and Refinement (IASR)

The scalability of augmentation was evaluated using GPT-3.5 Turbo, GPT-4 Turbo, GPT-4-o Mini, and Claude 3.5 Sonnet, across augmentation frequencies of 5x, 25x, 50x, 75x, and 100x paraphrases per sentence. Semantic preservation was assessed using cosine similarity previously shown schematically in Fig. 1. For Sentence 1, demonstrated in **Fig.4.a)** a syntactically simple imperative, all models maintained high semantic retention, with mean similarity ranging from 0.872 (Claude 3.5 Sonnet) to 0.8966 (GPT-4 Turbo), relative to their respective baseline similarities of 0.916, 0.937, and 0.932. GPT-3.5 Turbo exhibited the most consistent performance (mean: 0.8935; std: 0.0072–0.0233), while GPT-4o Mini reached its highest similarity of 0.8972 at 75x but displayed more variance. Claude 3.5 Sonnet, despite achieving a high score at 100x of 0.8860, showed greater inconsistency with increased augmentation volume.

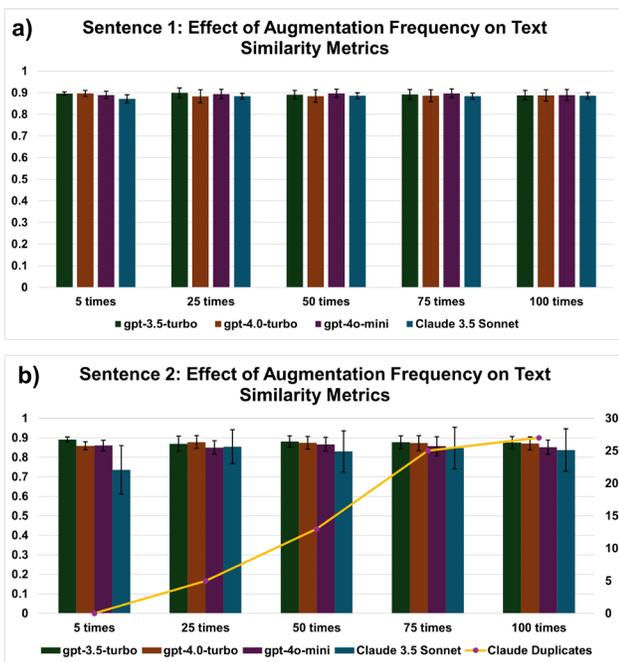

**Fig. 4.** Impact of augmentation frequency on semantic similarity across language models for two representative sentences. Cosine Similarity across augmentation frequencies scalability IASR workflow.

For Sentence 2, shown in **Fig.4.b)** which had a more complex syntactic structure, all models demonstrated a downward trend in semantic similarity as augmentation volume increased, relative to baseline values of 0.838 (GPT-3.5 Turbo), 0.859 (GPT-4 Turbo), 0.864 (GPT-4o Mini), and 0.942 (Claude). Claude 3.5 Sonnet experienced the steepest decline, from 0.8546 at 25x to 0.8372 at 100x, with standard deviation rising to 0.1093, indicating reduced reliability at higher scales. GPT-3.5 Turbo and GPT-4 Turbo remained comparatively stable, with GPT-3.5 achieving 0.8756 at 100x and maintaining a low standard deviation of 0.0312, highlighting its robustness under increasing paraphrasing pressure.

Redundancy analysis revealed significant duplication in

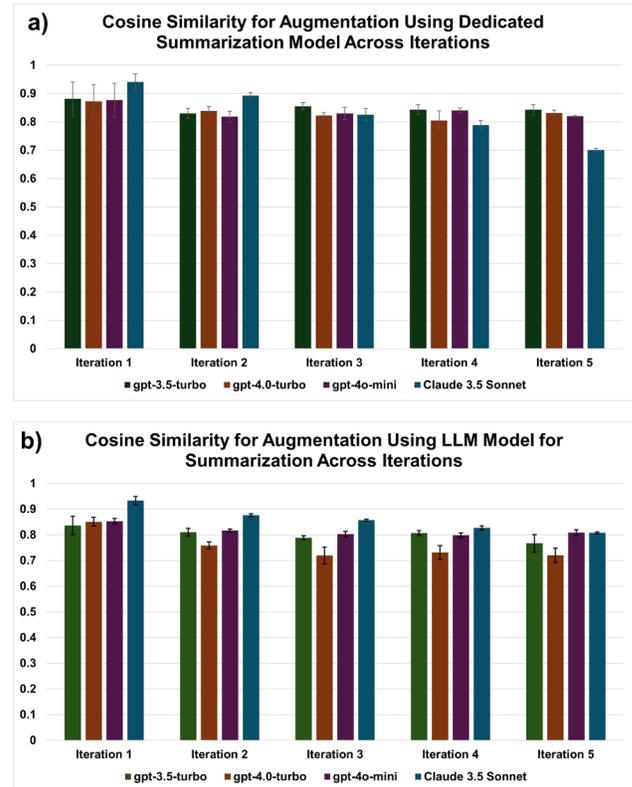

**Fig. 5**. Semantic stability across iterative augmentation cycles using the IASR framework with two summarization strategies: (a) a dedicated summarization model (T5-Base), and (b) integrated LLM-based summarization. Cosine similarity is used to evaluate semantic retention across five refinement iterations.

Claude 3.5 Sonnet's outputs, reaching a 27% duplication rate at 100x with 27 duplicates. At the same time, GPT-3.5 Turbo, GPT-4 Turbo, and GPT-4o Mini produced entirely novel paraphrases across all scales, suggesting stronger lexical diversity. In terms of efficiency, GPT-based models completed augmentation tasks in approximately 16 minutes, whereas Claude required 28 minutes, indicating a 75% increase in processing time. Combined with its duplication behavior and semantic inconsistency, this latency suggests Claude 3.5 Sonnet is less scalable for large augmentation workloads. Overall, GPT-3.5 Turbo and GPT-4o Mini demonstrated the strongest scalability, balancing semantic fidelity, diversity, and generation speed. Unlike the scalability test, which determined how much semantic fidelity models maintained as the volume of augmentation increased, the IASR framework was designed to assess whether iteratively repeated summarization cycles affect the semantic stability of augmented content. The results of T5-based and LLM-based IASR workflows are presented **Fig. 5.a)** and **Fig.5.b)** respectively. The repeated tests were designed to stress each model's resilience to repeated refinement, simulating real-world augmentation pipelines that require contextual holding across multiple cycles.

In the T5-based IASR setup, where summarization served as a diagnostic rather than a generative step, GPT-3.5 Turbo and GPT-4o Mini maintained high semantic retention across five iterations. GPT-3.5 Turbo achieved similarity scores between 0.8299 and 0.8812 (mean: 0.8503, std: 0.0205), while GPT-4o Mini scored 0.8184 to 0.8771 (mean: 0.8371, std: 0.0257). GPT-4 Turbo exhibited slightly more variation (mean: 0.8339, standard deviation: 0.0269), with a temporary dip observed in iteration 4. In contrast, Claude 3.5 Sonnet, despite a strong start at 0.9406, declined to 0.7008 by iteration 5, a 25.5% drop, suggesting it struggled to maintain alignment across refinement cycles. Although its mean standard deviation was moderate (0.1002), the steady decline suggested that Claude's generation style may conflict with the external summarization schema, leading to cumulative semantic drift.

In the LLM-based IASR setup, where the same model performed summarization, performance patterns shifted. GPT-3.5 Turbo again exhibited stability (mean: 0.8016, std: 0.0287), reaffirming its ability to preserve semantics through self-refinement. GPT-4 Turbo displayed greater drift (mean: 0.7559, std: 0.0654), with a marked decrease from 0.8506 to 0.7200 across iterations. GPT-4o Mini's similarity decreased modestly from 0.8531 to 0.8082, maintaining low variability (std: 0.0242). Claude 3.5 Sonnet performed more steadily here than in the T5 setup, falling from 0.9330 to 0.8072, with a respectable mean of 0.8601 and lower standard deviation (0.0506). These findings suggest Claude is better suited to self-summarization workflows, whereas T5-based external summarization exposes its limitations in maintaining contextual fidelity.

TABLE II
Comparative analysis of model performance across both the Scalability and IASR workflows.

| Metric | GPT-3.5 Turbo | GPT-4 Turbo | GPT-4o Mini | Claude 3.5 Sonnet |
|---|---|---|---|---|
| Scalability Avg (S1) | 0.893 | 0.887 | 0.893 | 0.882 |
| Scalability Avg (S2) | 0.878 | 0.871 | 0.857 | 0.821 |
| CV (S1) % | 2.15 | 2.87 | 2.33 | 1.64 |
| CV (S2) % | 3.31 | 3.58 | 4.27 | 13 |
| IASR T5 Avg | 0.85 | 0.833 | 0.837 | 0.829 |
| IASR LLM Avg | 0.801 | 0.755 | 0.815 | 0.86 |
| IASR T5 CV (%) | 2.93 | 3.06 | 2.63 | 2 |
| IASR LLM CV (%) | 2.52 | 3.1 | 1.16 | 0.89 |

TABLE II summarizes the mean cosine similarity, duplication rates, processing times, and coefficient of variation (CV). For scalability, measured via average cosine similarity for two representative sentences, S1 and S2, GPT-3.5 Turbo and GPT-4o Mini achieved the highest average semantic retention of 0.893 for S1, with Claude 3.5 Sonnet scoring slightly lower at 0.882. However, for the more syntactically complex sentence (S2), GPT-3.5 Turbo again led around 0.878, while Claude 3.5 Sonnet exhibited the lowest score of 0.821, indicating reduced robustness under increased augmentation frequency. Coefficient of variation (CV) further revealed model stability; GPT-3.5 Turbo and GPT-4 Turbo maintained low variability for both S1 and S2, whereas Claude 3.5 Sonnet showed a sharp increase in variability for S2 at CV = 13%, suggesting semantic drift under scale. In the IASR setting, GPT-3.5 Turbo achieved the highest average similarity across both summarization strategies of 0.85 (T5-based) and 0.801 (LLM-based) with moderate variation of CV ≈ 2.5–2.9%. Claude 3.5 Sonnet, although achieving the highest LLM-based similarity of 0.86, exhibited notable inconsistency across iterations in the scalability setup. Overall, GPT-3.5 Turbo demonstrated the most stable and semantically faithful performance across both augmentation and summarization-intensive pipelines. These findings collectively indicated that while outside summarization by T5 enables gradual improvement for certain models, self-improvement by LLMs can lead to more consistent outcomes for others. GPT-3.5 Turbo was the most consistent of both versions, whereas Claude 3.5 Sonnet became increasingly coherent when executed within a self-summarization cycle.

*B. Topic modeling Evaluation with and without LLM-based Augmentation*

After analyzing the augmentation outcomes, GPT-3.5 Turbo was selected to generate augmented data due to its high semantic retention, computational speed, and affordability. To assess the downstream impact of augmentation on topic modeling performance, four configurations of BERTopic and GPT-based topic labeling were evaluated. Each of the original responses was augmented three times to enhance dataset size without altering contextual meaning. Topic modeling was then conducted using BERTopic under four experimental conditions to isolate the individual and joint effects of augmentation and GPT-based topic labeling. These were: (1) BERTopic on Non-Augmented Data, (2) BERTopic on Augmented Data, (3) BERTopic with few-shot GPT-based labeling on Non-Augmented Data, and (4) BERTopic with few-shot GPT-based labeling on Augmented Data. This setup provided a structured ablation study that assessed the contributions of augmentation and structured topic refinement against key performance metrics, including topic coherence, diversity, distinctiveness, and internal consistency.

Table III summarizes the performance of various topic modeling configurations evaluated on coherence, diversity, overlap, and structural metrics. The baseline BERTopic model, applied to unaugmented data, yielded low topic coherence of 0.469 , low intra-topic similarity of 0.454, and extracted only 5 topics indicating poor topic granularity and specificity. Applying BERTopic to augmented data increased the topic count to 28 and improved both coherence of 0.505 and intra-topic similarity of 0.615, but also introduced redundancy, as reflected in a consistent topic overlap of 0.041. When GPT-based clustering was applied without data augmentation, the number of topics remained low at 4, but the output showed improved coherence of 0.510, perfect topic diversity of 1.0, and zero overlap highlighting GPT's capacity for enhancing topic structure. The best overall performance was achieved using BERTopic + GPT on augmented data, which balanced topic granularity (20 topics), coherence of 0.526, intra-topic similarity of 0.518, and inter-topic distance at 0.650, all while eliminating topic overlap. These results demonstrated that LLM-based augmentation paired with GPT-guided clustering yields more coherent, diverse, and semantically distinct topic structures compared to traditional unsupervised pipelines.

TABLE III
Ablation Study on the Impact of Augmentation and GPT-Assisted Topic Extraction on Topic Modeling

| Metric | Baseline (BERTopic Only) | BERTopic on Augmented Data | BERTopic + GPT on Non-Augmented Data | BERTopic + GPT on Augmented Data |
|---|---|---|---|---|
| Topic Coherence | 0.469 | 0.505 | 0.510 | 0.526 |
| Topic Diversity | 0.64 | 0.851 | 1 | 1 |
| Topic Overlap | 0.041 | 0.041 | 0 | 0 |
| Number of Topics | 5 | 28 | 4 | 20 |
| Intra-Topic Simialrity | 0.454 | 0.615 | 0.491 | 0.518 |
| Inter-Topic Distance | 0.307 | 0.724 | 0.441 | 0.650 |
| Observations | Limited topic granularity and coherence; extracted topics lack specificity. | Augmentation increased the number of topics but introduced redundancy and overlap. | GPT improved topic distinctiveness and structure but extracted fewer topics. | Best overall performance with optimal topic granularity, coherence, and distinctiveness. |

When GPT-assisted extraction was applied without augmentation, BERTopic + GPT on Non-Augmented Data, topic coherence improved slightly (0.5101 vs. 0.4697), and redundancy was eliminated with a topic overlap of 0.000.

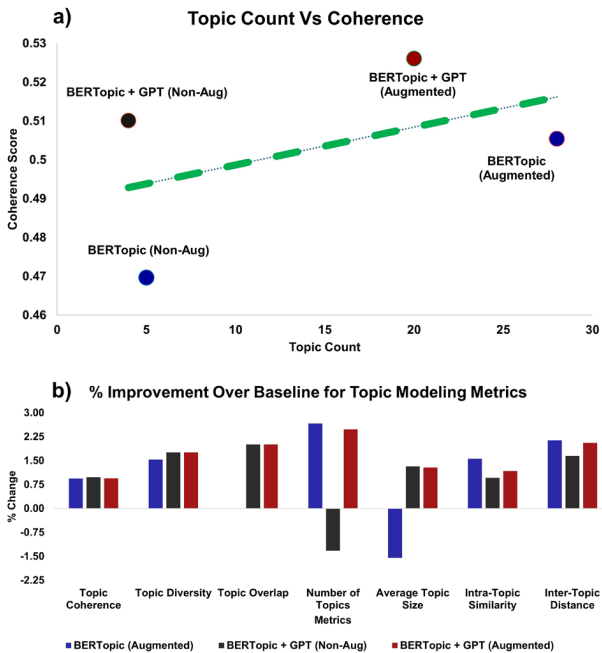

Fig. 6. Evaluation of topic modeling performance across configurations with and without data augmentation and GPT-enhanced clustering a) Topic Count vs Coherence, (b) % Improvement Over Baseline for Topic Modeling Metrics [transformed as sign (x)· 〚log〛_10 (1+|x|)]

However, this configuration extracted only four broad topics, like "Faculty Support & Compensation", "Workloads, Faculty Morale, Recognition", and "Faculty Behavior and Accountability". This suggests that GPT's structured extraction process helps refine topic definitions but is insufficient in producing diverse and granular topic clusters without augmentation. The most effective configuration was BERTopic + GPT + few-shot learning on Augmented Data, which optimized all key metrics. This setup extracted 20 well-separated topics, summarized in TABLE IV, such as "Administration: Transparency & Faculty Support", "Compensation Disparities in Academia", and "Faculty Support: Non-STEM Engagement".

TABLE IV
Extracted Topics from Augmented Dataset

| Sr.no. | Extracted Topic | Count |
|---|---|---|
| 1 | Administration: Transparency & Faculty Support | 212 |
| 2 | Compensation Disparities in Academia | 87 |
| 3 | Workload: Faculty Overburden & Support | 41 |
| 4 | Faculty Compensation & Administration Issues | 36 |
| 5 | Parental Leave Policy: Dual-Employed Faculty | 25 |
| 6 | Leadership Compensation & Campus Integration | 25 |
| 7 | Age Discrimination | 24 |
| 8 | Faculty Support: Non-STEM Engagement | 23 |
| 9 | Work Environment: Discrimination & Faculty Support | 21 |
| 10 | Strategic Planning: Educational Prioritization | 21 |
| 11 | Survey Design: Minimizing Bias | 20 |
| 12 | Leadership Accountability & Transparency | 16 |
| 13 | Workload: Faculty Productivity & Work-Life Balance | 16 |

| 14 | Culture: Accountability & Misconduct | 13 |
| --- | --- | --- |
| 15 | Regulatory Compliance: Efficiency & Oversight | 12 |
| 16 | Childcare Policies & Support | 12 |
| 17 | Leadership: Transparency & Healthcare Enhancement | 12 |
| 18 | Administration: Support & Research Discrimination | 12 |
| 19 | Work-Life Balance: Humanities vs. Ohio Legislature | 12 |
| 20 | Faculty Retention & Workload Equity | 11 |
| Topics reflect patterns from augmented faculty responses, illustrating the method's ability to capture thematic structures in open-ended text. | | |

As shown in **Fig. 6.(a)**, the augmented BERTopic + GPT pipeline achieved the best trade-off between topic granularity and coherence, generating 25 coherent and distinct topics with a coherence score of 0.5261, outperforming all other variants. The non-augmented BERTopic baseline, in contrast, identifies fewer and less coherent topics. The augmented configuration also showed strong structural separation of inter-topic distance: 0.6507, high intra-topic consistency with similarity: 0.5183, and zero topic overlap indicating both distinct and internally cohesive clusters. These results confirmed that augmentation enhances dataset richness and increases topic granularity, while GPT-assisted extraction plays a crucial role in structuring and refining the extracted topics for better interpretability.

In addition, statistical analysis supported the observed improvements in topic modeling with augmentation. Paired t-tests between non-augmented and augmented environments, keeping GPT-based representation constant, showed a sharp improvement in inter-topic distance, p = 0.0194, d = –4.0871, which infers better topic separation after augmentation as shown in TABLE V. Though coherence wasn't statistically significant, p = 0.2422, the large effect size, d = 0.9484, suggests practical improvements. Similarly, the reduced average topic size reflected more compressed topics; however, the change did not reach a significant value, p = 0.2917. These results, cumulatively, validated prior statements of improved structure and granularity from augmentation. This is further illustrated in **Fig. 6.(b)**, which presents percentage improvements across key topic modeling metrics, benchmarked against the baseline BERTopic under on-augmented condition. Augmented pipelines particularly BERTopic + GPT demonstrate consistent gains across topic modeling metrics including topic coherence, diversity, overlap, and average topic size. Improvements range from ~0.75% to over 3% across metrics, confirming that LLM-based augmentation enhanced semantic coverage and cluster quality.

TABLE V
Paired t-test Results for Topic Modeling Metrics (Augmented vs. Non-Augmented)

| Metric | Coherence | Distance | Topic Size |
| --- | --- | --- | --- |
| **Mean (Non-Aug)** | 0.552 | 0.346 | 44.65 |
| **Mean (Aug)** | 0.529 | 0.6267 | 33.299 |
| **T-Stat** | 1.642 | -7.079 | 1.419 |
| **P-Value** | 0.242 | 0.019 | 0.291 |
| **Cohen's d** | 0.948 | -4.087 | 0.819 |

Overall, the ablation study strengthened the complementary nature of augmentation and GPT-assisted topic extraction. Augmentation alone expanded topic coverage but can introduce redundancy and fragmentation, whereas GPT-assisted topic extraction alone improves coherence and topic differentiation but lacks the diversity needed for comprehensive analysis. A combined approach achieved the best balance between granularity, coherence, and distinctiveness, ensuring semantically meaningful and interpretable topic distributions. These findings accentuate the synergy of augmentation and GPT-based theme labeling. While augmentation on its own amplified topical diversity, at times it injected redundant semantics and structure incoherence. GPT-based labeling augmented thematic coherence and semantic invariance but lacked topical diversity when applied singly without augmentation. Thus, the hybrid strategy created the most understandable and semantically distinctive themes with the best-balanced improvement across diversity, coherence, and structure consistency. These insights were particularly valuable for analyzing unstructured textual data, where maintaining both broad thematic coverage and precise topic differentiation is essential for extracting actionable insights across various domains.

V. Conclusion and Future Work

This study introduced two evaluation frameworks—Scalability Analysis and Iterative Augmentation with Summarization Refinement (IASR) to assess the semantic fidelity and downstream utility of LLM-generated text augmentation under data-scarce conditions. Unlike conventional rule-based or lexical substitution approaches, the proposed frameworks emphasize semantic-level evaluation, incorporating cosine similarity metrics, duplication tracking, and topic modeling performance to offer a principled assessment of augmentation quality. Among the evaluated models, GPT-3.5 Turbo demonstrated superior performance, yielding high similarity scores across varying augmentation sizes, zero duplicate generations, and minimal semantic drift across iterative summarization cycles. These findings underscore the model-dependent nature of augmentation quality and highlight the importance of using iterative, context-aware evaluation pipelines for NLP tasks. To validate the downstream utility of the augmented data, topic modeling was conducted using BERTopic with and without GPT-enhanced labeling. The augmented datasets, when combined with GPT-based few-shot topic labeling, led to substantial improvements in topic coherence, diversity, and separation. The optimal configuration achieved a 400% increase in topic count relative to the baseline, while eliminating redundancy and preserving semantic structure. These results are particularly impactful for real-world applications involving short-form, domain-specific texts such as survey responses, clinical notes, or incident reports. Despite these gains, the augmentation process lacked explicit control mechanisms, relying instead on template-driven prompting and post-hoc filtering via semantic similarity. This limitation may affect reproducibility and interpretability across deployments. Future work can address this by incorporating prompt attribution models, domain-adaptive fine-tuning, or integration with open-weight LLMs. Exploring encoder-based contextual models like MiniLM and BERT, or hybrid encoder-decoder architectures such as T5 and PEGASUS, may further balance generative flexibility with semantic stability. Expanding the IASR framework to include transformations like back-translation, style transfer, or sentiment modulation could deepen insights into augmentation robustness. As LLMs evolve, their ability to enhance sparse data while supporting structured, interpretable analysis will be central to advancing reliable and scalable NLP systems.

Code Availability: The code used in this research is accessible at https://github.com/anjurgupta/IASR-Survey-Data

Ethical Considerations and Data Privacy Statement: This study involved the secondary analysis of de-identified responses originally collected through an institutional organizational climate survey. All data used in this research were stripped of personally identifiable information prior to analysis. The original data collection was approved by the Institutional Review Board (IRB) under a protocol 301865- UT focused on organizational assessment and did not explicitly include provisions for artificial intelligence (AI) or machine learning (ML) analyses. To mitigate potential risks associated with re-use of human-subject data, the current analysis was performed exclusively on anonymized text, and no attempts were made to infer or link responses to individuals or demographic identifiers. The AI-based augmentation and evaluation procedures were implemented with a focus on aggregate trends and model behavior rather than individual-level profiling.

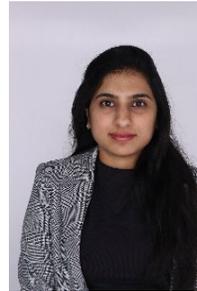

**Payal Bhattad** received the BE degree in Information Technology from Yeshwantrao Chavan College of Engineering, Nagpur, India, in 2016 and the MS degree in Computer Science from The University of Toledo, Toledo, OH, USA, in 2025. She worked as a software engineer at Tata Consultancy Services, India. Her current research interests include natural language processing.

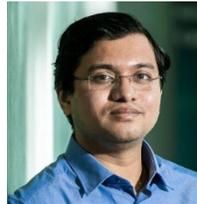

**Sai Manoj Pudukotai Dinakarrao, (Senior IEEE Member)** an assistant professor in the Department of Electrical and Computer Engineering at Mason, Pudukotai Dinakarrao's research interests include hardware security, adversarial machine learning, Internet of Things networks, deep learning in resource-constrained environments, in-memory computing, accelerator design, algorithms, design of self-aware many-core microprocessors, and resource management in many-core microprocessors. He was a recipient of The Young Research Fellow Award in the Design Automation Conference (DAC) 2013.

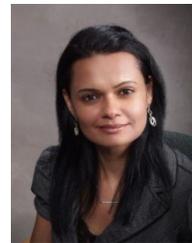

**Anju Gupta** is an Associate Professor in the Department of Mechanical Engineering at the University of Toledo, Ohio, USA. Her interdisciplinary research spans interfacial thermal-fluids, and data-driven modeling, with a growing emphasis on applying machine learning to complex physical and social science systems.